\documentclass[10pt,twocolumn,letterpaper]{article}
\usepackage{iccv}
\usepackage{times}
\usepackage{epsfig}
\usepackage{graphicx}
\usepackage{amsmath}
\usepackage{amssymb}
\usepackage{multirow}
\usepackage{enumitem}
\DeclareMathOperator{\softmax}{softmax}
\DeclareMathOperator{\Attention}{Attention}
\DeclareMathOperator{\MultiHead}{MultiHead}
\DeclareMathOperator{\concat}{concat}
\DeclareMathOperator{\head}{head}

\newcommand{\newparagraph}[1]{{\vspace{2pt}\noindent\textbf{#1}~}} 

\usepackage[pagebackref=true,breaklinks=true,letterpaper=true,colorlinks,bookmarks=false]{hyperref}

\def\iccvPaperID{5165} 

\iccvfinalcopy
\ificcvfinal \fi
\begin{document}

\title{Fine-Grained Object Detection \\ over Scientific Document Images with Region Embeddings}

\author{Ankur Goswami \thanks{Equal Contribution}\\
University of Wisconsin, Madison\\
{\tt\small ankurg@cs.wisc.edu}
\and
Joshua McGrath \footnotemark[1]\\
University of Waterloo\\
{\tt\small jmcgrath@waterloo.ca}
\and
Shanan Peters\\
University of Wisconsin, Madison\\
{\tt\small peters@geology.wisc.edu}
\and
Theodoros Rekatsinas\\
University of Wisconsin, Madison\\
{\tt\small thodrek@cs.wisc.edu}
}

\maketitle
\begin{abstract}
	We study the problem of object detection over scanned images of scientific documents. We consider images that contain objects of varying aspect ratios and sizes and range from coarse elements such as tables and figures to fine elements such as equations and section headers. We find that current object detectors fail to produce properly localized region proposals over such page objects. We revisit the original R-CNN model and present a method for generating fine-grained proposals over document elements. We also present a region embedding model that uses the convolutional maps of a proposal's neighbors as context to produce an embedding for each proposal. This region embedding is able to capture the semantic relationships between a target region and its surrounding context. Our end-to-end model produces an embedding for each proposal, then classifies each proposal by using a multi-head attention model that attends to the most important neighbors of a proposal. To evaluate our model, we collect and annotate a dataset of publications from heterogeneous journals. We show that our model, referred to as Attentive-RCNN, yields a 17\% mAP improvement compared to standard object detection models.\end{abstract}

\section{Introduction}\label{sec:intro}
Much of the scientific knowledge is encoded in documents in the form of tables, figures, sets of coupled equations, etc. Access to this information can dramatically speed up the pace and consistency of scientific discoveries and their application to urgent problems. Several projects~\cite{Ammar2018ConstructionOT, zhang_geodeepdive:_2013} aim to make this data more accessible by building central repositories of scientific papers that researchers can use to access information that might be hidden away in tables, equations, and figures within the papers. Many of these methods rely on optical character recognition (OCR) engines such as Tesseract~\cite{smith_overview_2007} to extract useful information from the aforementioned elements. However, applying an OCR model meant for English text extraction to an image of a formula that contains symbols outside the English language will yield an unsatisfactory output. A natural need emerges to classify different parts of the document image, a task known as page object detection.

\begin{figure}[t]
\begin{center}
  \includegraphics[width=1.0\linewidth]{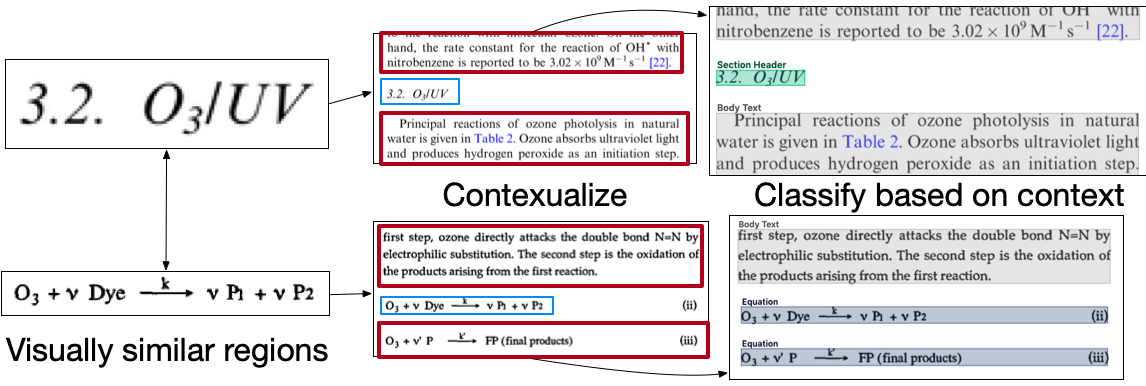}
\end{center}
   \caption{Contextualization is needed to differentiate between similar object classes.}
\label{fig:fig1}
\end{figure}

Similar to object detection, page object detection requires solving two complex tasks. The detector must solve: (1) a {\em recognition problem}, i.e., distinguish foreground objects from background objects and assign them the proper object class labels, and (2) a {\em localization problem}, to assign accurate bounding boxes to different objects.  In the case of scientific documents, these tasks become challenging due to how varied different document layouts are---it is common to observe significant variability of aspect ratios and sizes of objects (even within the same class) across different documents---and how difficult it is to define features that a classifier can use to differentiate between visually similar structures. For example, a line of English text (e.g., a section header) and a mathematical formula can be hard to differentiate because formulas are often composed of English letters as in.

To address these challenges recent information extraction pipelines~\cite{staar_corpus_2018, yi_cnn_2017} attempt to transfer the success of the R-CNN model in performing object detection in 3D scene images to the domain of document images. These works explore the use of state-of-the-art R-CNN models such as Faster-RCNN~\cite{NIPS2015_5638}, which cast object detection as a multi-task learning problem that combines classification and bounding box regression over {\em automatically proposed regions} of interest. However, they identify that such methods fail to perform accurate location in the document space, a finding that is consistent~\cite{DBLP:conf/cvpr/CaiV18} with benchmark image domains such as COCO. To address this issue, current page object detection pipelines either rely on bespoke, error-prone ensembles of detection techniques that are tailored to specific object classes~\cite{Ammar2018ConstructionOT, staar_corpus_2018, DBLP:conf/cvpr/YangYAKKG17} or combine algorithmic page segmentation to propose regions with a CNN-based classifier, in line with the original R-CNN model~\cite{yi_cnn_2017}. A fundamental flaw with the original R-CNN, however, is that each proposal is classified independently from all other proposals, which means whatever information is outside the proposal cannot be used for classification. Contextual information, however, is often key to differentiate between objects that are fundamentally similar (consider the aforementioned comparison between text and formula shown in Figure~\ref{fig:fig1}).

Inspired by recent advances in language modeling, we propose a framework that creates region embeddings: a learned representation for regions trained on the feature maps of proposals and proposal neighbors. After embedding each proposal region, we then are able to utilize multi-head attention to learn which of a proposal's neighbors to attend to for classification. Analogous to language modeling, proposals must be sufficiently independent such that it is clear which neighbors contain contextual, non-redundant information, similar to how words in a sentence are separated by spaces. We propose a grid layout segmentation algorithm for documents that divides each page into a grid, and then refines each cell in the grid to fit the content. Naturally, each cell in the grid is spatially independent from all other cells, and a cell's neighbors are clear. In our post-processing, cells are merged with their neighbors based on their class to create highly localized results. We call our end to end model Attentive-RCNN. Our core contributions within Attentive-RCNN are as follows:

\begin{enumerate}[noitemsep,topsep=0pt]
    \item We propose a simple and effective grid page segmentation algorithm that recalls more objects with fewer proposals than comparative methods.
    \item We utilize a region embedding model that generates embeddings for each proposal region, and classifies them using a multi-head attention classification head.
    \item We provide an open source implementation of our page object detection system~\footnote{https://github.com/UW-COSMOS/Cosmos}.
\end{enumerate}

To verify our results, we annotate a dataset of 2,466 scientific document pages, considering object classes that contain tables, figures, equations, body text, references, section headers, etc. These documents are composed of scientific papers that date back as early as 1970. The layout of these documents as well as the quality of the scans are highly variable~\footnote{We aim to make this dataset publicly available to encourage further work on this problem.}. We present experimental results showcasing our system performing better than competing page object detection systems for standard object detection metrics. Our model produces a $+17\%$ mAP and $+12$  points information retrieval F1 improvement over a Faster-RCNN baseline. Compared to a standard fully convolutional network (FCN) classification head that also utilizes our proposals, our attention model produces a $+3\%$ mAP and $+4$ information retrieval F1 improvement. To show how our results generalize to other document sets, we also present our system's results on the ICDAR POD 2017 competition dataset, which is a similar page object detection task to the one we discuss, but with only table, figure, and equation classes. We show that our model is able to retrieve 1.15 times as many valid objects compared to the FCN baseline.

\begin{figure*}
\begin{center}
	\includegraphics[width=1.0\linewidth]{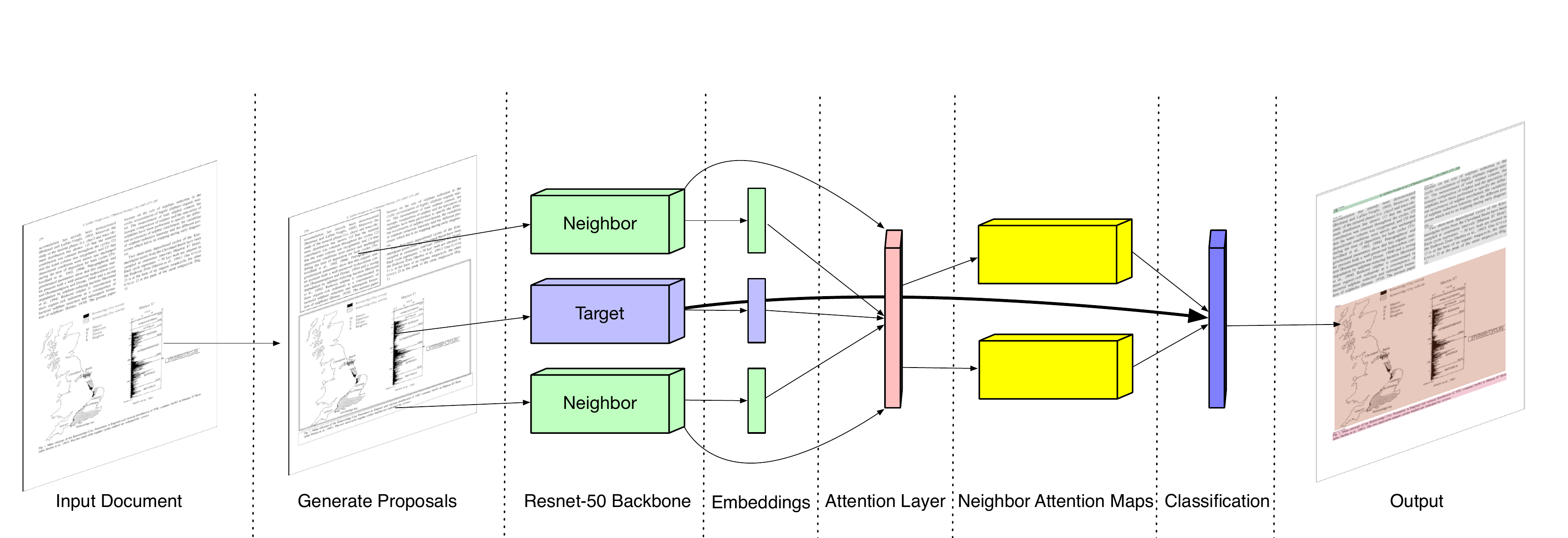}
\end{center}
 \caption{Attentive R-CNN end to end architecture.}
\label{fig:architecture}
\end{figure*}

\section{Related Work}\label{sec:related}
This work primarily builds on attempts to apply object detection models to the page object detection task, as well as models that incorporate contextual features into standard object detection. Our application of attention to the context of object classification is directly inspired by the success of attention mechanisms in language modeling~\cite{DBLP:journals/corr/BahdanauCB14, NIPS2017_7181}.

The work on the page object detection task most similar to ours is that of Yi et al.~\cite{yi_cnn_2017}. Unlike other work on this task, they address text in addition to tables, figures, and equations by collecting a text line class in their dataset. Their model, however, uses algorithmic proposals that they feed to an AlexNet head, which is not significantly different from doing architecture search over the original R-CNN, with some bells and whistles to handle overlapping proposals. We differentiate from R-CNN through our attention mechanism, where we are fully taking advantage of how much easier it is to segment 2D scientific documents compared to 3D scene images. Our grid proposal algorithm is inspired by connected component based document segmentation algorithms such as \cite{chen_hybrid_2013}.

The ICDAR 2017 Page Object Detection challenge \cite{gao_icdar2017_2017} is a recent competition that focused on only detecting tables, figures, and equations. Models that participated in that competition primarily tried to adapt Faster-RCNN to the problem, to varying degrees of success. Though a fundamental part of our motivation is identifying all objects on a page, and our proposal algorithm and model reflect this, we evaluate our model on this dataset. Others build models specifically to extract tables \cite{clark_extracting_nodate, gilani_table_2017, schreiber_deepdesrt:_2017, shigarov_configurable_2016, hao_table_2016, 6065417}, figures \cite{he_multi-scale_2017}, and equations \cite{gao_deep_2017, Lin2014, toumit_hierarchical_1999, drake_distinguishing_2005}. Because the state of the art on this task is based on R-CNN architectures, we compare to R-CNN based architectures in our experiments.

Visual Attention Models have been explored in~\cite{ba2014multiple, Mnih:2014:RMV:2969033.2969073}, however, both these works impose sequential relations between objects in an image in order to make use of context in a recurrent neural network. In this work, we learn to attend to a bag of inputs based on a learned attention model. Spatial Transformer Networks are another form of visual attention 
mechanism explored in \cite{NIPS2015_5854}. Spatial Transformer Networks allow for manipulation of convolutional maps as a form of learned self-attention. The goal of Attentive-RCNN is to use context from neighboring regions to make predictions about a target region, and so self attention is not explored.

\section{Attentive-RCNN}\label{sec:attrcnn}
We now introduce the Attentive R-CNN architecture. An overview diagram is shown in Figure~\ref{fig:architecture}. As shown, our model is composed of three modules: (1) A proposal algorithm which leverages domain knowledge to generate high quality regions of interest for each input document image; (2) An attention model which provides context-aware information to the classification subnetwork; (3) A classification model which takes into account the current region as well as its context to make predictions over the classes of interest. We describe each module in the subsequent sections.

\subsection{Grid Proposal Algorithm}\label{sec:gridprop}
The first step of Attentive-RCNN is to produce a collection of region proposal over an input document image. Scientific papers, regardless of origin or journal, generally adhere to some regular grid structure. Finding proposals over the document, then, is finding an appropriately sized grid. Our algorithm follows the three steps described below.

\newparagraph{Step 1:} We preprocess each image for proposal generation by converting it to a binary map. We balance the margins of the binary map by checking the distance from each side of the page to the nearest pixel. If a side is shorter by distance $d$, we remove $d$ pixel columns from the opposite side. 

\newparagraph{Step 2:} We next divide each map into rows. For some row height $R$, we find all blank rows of height at least $R$. For our experiments, we set $R$ to 15 pixels. For each row between the found blank rows, we now detect whether the row can be divided into 1-3 columns. For each column number $c$, find $c-1$ division lines that divide the row into $c$ columns. If a division line intersects with a non-white pixel, we check the next column number. For all valid divisions, we take the maximum column number and split the row into that number columns. For each row-column cell, we divide the cell into rows again using the same row division process.

\newparagraph{Step 3:} Finally, we refine each cell by finding all 8-connected components over the grid defined by the cell, filtering connected components that are too small, then taking the minimum bounding box encapsulating all connected components that remain.

Because of discrepancies between page layouts, $R$ is sometimes too low and overzealously divides cells. As a simple way to mitigate dividing a page into page lines, we check the average height of all final rows. If the average height is less than $3 \cdot R$, we redivide the page, doubling $R$ for the new iteration. Our final result is a grid of proposals, each ensured not to have any overlap from another, but still approximately containing a page element. Objects that naturally contain white space to separate their content, such as tables, can be decomposed into sub-objects by this algorithm. If these sub-objects are successfully classified as their overall object class, neighboring cells are merged together based on their class at a post-processing step (see Section~\ref{sec:postprocessing}).

\subsection{Attentive Neural Network}\label{sec:attnn}
Given the region proposals output by the algorithm in Section~\ref{sec:gridprop}, Attentive-RCNN is used to perform classification. At a high-level Attentive-RCNN follows the steps shown in Figure~\ref{fig:architecture}: Attentive-RCNN takes as input a target region proposal together with its context (corresponding to its neighbor regions) and passes them through a ResNet50 backbone to obtain their convolutional maps. To classify the target region, Attentive-RCNN takes the convolutional maps of the region and its context and feeds those through an embedding and attention mechanism to produce context convolutional maps. Finally, the generated context maps and the target region convolutional map are fed to a classification head. We discuss each of these steps in detail.

\newparagraph{Region Embeddings}
To use an attention model over the target region proposal and its context (i.e., its neighbor proposals as defined in the next paragraph) we need a method to determine the relevance between the corresponding sub images. We draw inspiration from the skip-gram embeddings in natural language processing~\cite{Mikolov:2013:DRW:2999792.2999959}. Given a target proposal and its neighboring proposal, we use a 2 layer Neural Network to embed each of the corresponding image into a $d_k$-dimensional vector space. The embeddings are pre-trained using a negative sampling technique first explored in~\cite{Mikolov:2013:DRW:2999792.2999959}. We form the set of training examples as follows: given a target region proposal, we find its neighboring region proposals and use those to form positive training examples. We construct negative training examples by sampling non-neighboring region proposals from the same document image. We then fit the distribution $P(D = 1| Target, Neighbor)$ where $D=1$ if the target-neighbor pair is a positive pair, and $0$ if the target-neighbor pair correspond to a negative pair. Given two embeddings from the Neural Network $v_{c}$ and $v_{n}$ we  predict $P(D=1| Target, Neighbor) = \frac{1}{1 + e^{-v_{c} \cdot v_{n}}}$. This is the formulation used to produce skip-gram embeddings but adopted over convolutional maps of images corresponding to region proposals.

\newparagraph{Visual Attention}
To define visual attention we are given a document image, a set of region proposals, and an index to the  target region proposal that we want to classify. We first identify the neighboring region proposals that form the context of the target proposal. We find the neighbors of a target region proposal as follows: We expand the bounding box of the target region by $\delta_{neighbor}$ pixels in each direction. Using this new bounding box, we check for overlaps with other region proposals,  and any region with nonzero IoU with this new bounding box is defined as a neighbor of the target region proposal. Figure\ref{fig:attentionexample} illustrates this process.

\begin{figure}
\begin{center}
	\includegraphics[width=1.0\columnwidth]{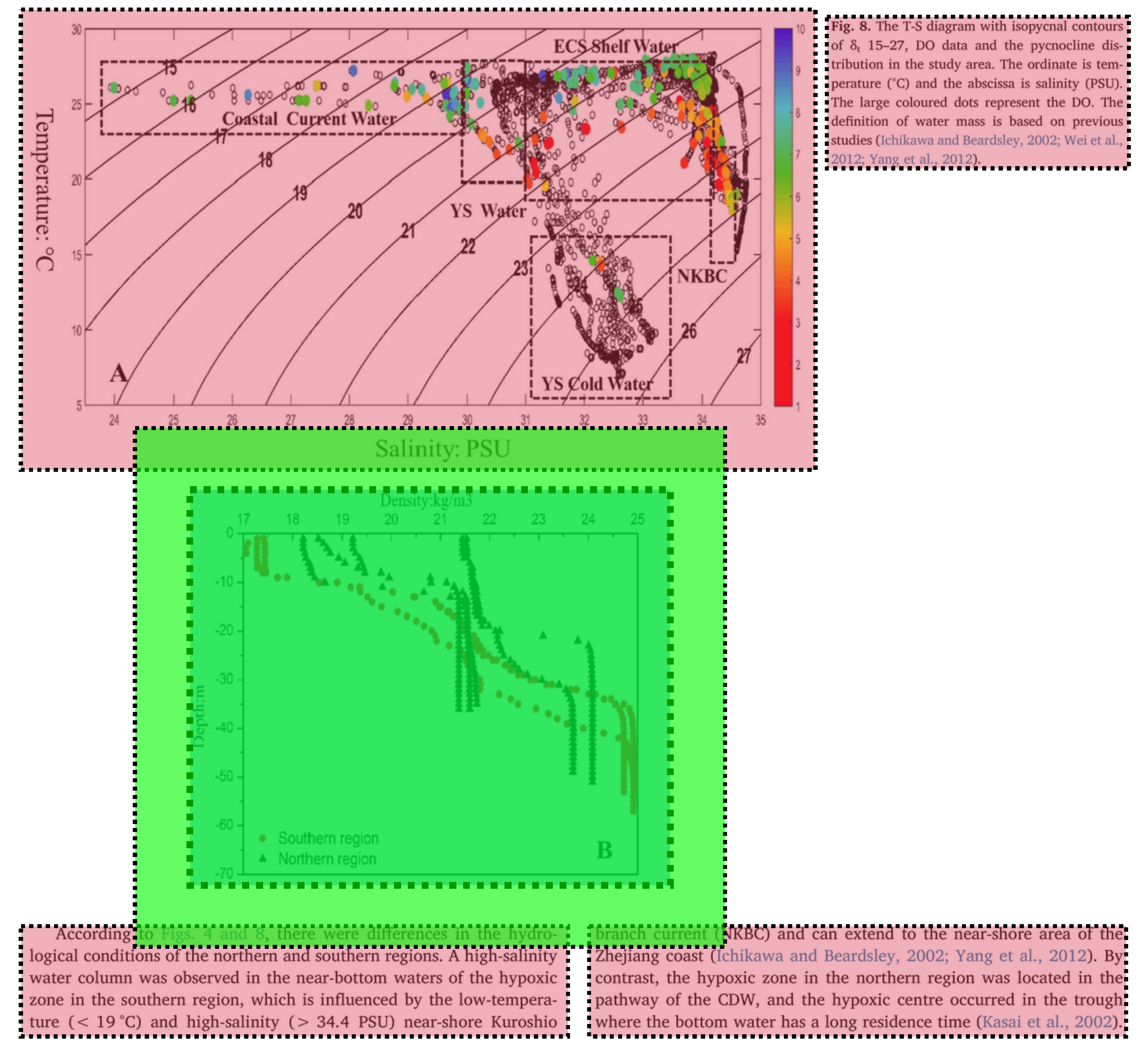}
\end{center}
  \caption{An example of the neighbor finding process. The inner of the blue/green boxes is the target region, and the red regions are potential neighbors. The outer green box represents the expansion delta. Any pink box which intersects the out green box is considered a neighbor of the target.}
\label{fig:attentionexmaple}
\end{figure}

Given a target proposal and its context, we embed each region and feed them into a multi-head attention model as seen in~\cite{NIPS2017_7181}. Given the target region embedding $q$, a set of neighbor embeddings $K =\{k_1, k_2, \ldots k_m\}$  and a corresponding set of convolutional maps $V = \{V_1, V_2, \ldots V_m\}$ the scaled dot product attention is defined as:

$$\Attention(q, K, V) = \sum_{i =1 }^m \softmax\left(\frac{qK^T}{\sqrt{d_k}}\right)_i V_i$$

Given region embeddings $q$ and $k_i$---where $k_i$ is a neighbor of the target region---their dot product should have a large magnitude and saturate the softmax if $k_i$ is informative for predicting the class of $q$. This means that region proposal corresponding to $k_i$ should be attended to. In order to attend to multiple context regions at once, Attentive-RCNN adopts a multi-head attention model. Each head is based off a learned projection of the original embeddings. Formally, multi-head attention is defined by:

\begin{multline*}
  \MultiHead(q,K,V) = \concat\left(\head_1,\ldots, \head_n \right)
  \\\text{where} \head_i =\Attention(W_i^{(q)}q,W_i^{(k)}K,V)
\end{multline*}
where each $W_i$ is a matrix of learned parameters, this allows Attentive-RCNN to attend to different kinds of inputs based on their relevance to the classification
task.

\paragraph{Classification} Given the convolutional feature map of a target region proposal, and the feature maps produced by the aforementioned attention model, a three layer Neural Network combines these feature maps and classifies the current region using a softmax layer. 

\subsection{Training the Attentive-RCNN Model}
The training procedure for Attentive-RCNN happens in two steps. First, the embeddings are pre-trained on a Noise-Contrastive Estimation \cite{gutmann2010noise} objective that comes from natural language processing techniques. Pretraining of embeddings allows the model to converge more consistently, and is similar to Faster-RCNN's mutli-step training scheme from \cite{NIPS2015_5638}. After the embeddings are trained, the embeddings, attention model and classification modules are trained end-to-end on a multi-class classification objective (computed via the cross-entropy loss). For regularization we use Weight Decay, and Dropout at each layer of the Attention, Classification, and Embedding networks. The ResNet50 backbone uses Batch Normalization during training time.

In order for the system to learn the actual distribution of images that it will receive at inference time, it is given regions that come from the proposals algorithm as opposed to the ground truth. To assign ground truth targets to each proposal we find the maximum-overlap ground truth region for each proposal and assign it to that ground truth class. Proposals with no overlap are filtered out.

\newparagraph{Implementation Details}
For embeddings, we use 256-d vectors for each region. This is not a carefully chosen dimension, but instead a usual value for embeddings in an NLP context. As our experiments show, this dimensionality still allows our model to achieve good results. We train a variant of Attentive-RCNN with three attention heads. This is chosen for the maximum number of neighbors in a two-column paper context. In a general object detection regime, more heads may be preferred. At training time, we use a weight decay of $10^{-5}$ and a learning rate of $3*10^{-4}$ with the Adam optimization algorithm~\cite{DBLP:journals/corr/KingmaB14}. The dropout probability is $0.5$. We use a batch size of six.

\subsection{Postprocessing}\label{sec:postprocessing}
After final classification, we further refine our results via a post-processing step.
First, figure and table captions are easy to classify if the text within a proposal is available. Because we assume that pdf metadata is not necessarily available, we run Tesseract OCR over each proposal. We first focus on handling possible confusions between Body Text and Figure Caption or Table Caption. We consider proposals predicted as Body Text and consider the first token of the OCR output. If the first token contains `fig' or `figure', we change it to Figure Caption. Similarly for Table Captions, if the first token of the OCR output is a token that matches to `table', `tab' or `tbl' we change the target proposal classification to Table Caption.

Second, for certain document layouts our grid-based proposal generation algorithm can decompose objects whose content is separated by whitespace (e.g., tables without grid lines) into sub-objects. We merge divided sub-figures into a single figure by following the assumption that there exists a one-to-one mapping between figures and figure captions. A similar rule is used for tables.

To obtain a merged table or figure, we combine neighboring sub-figures and sub-tables into a single figure and table. For each pair of figures or tables on a page, we take a bounding box over their union. If the union does not overlap with any other objects on the page, we replace the individual bounding boxes with their union bounding box. For tables, we more aggressively merge, allowing merges that overlap with any class except body text or captions. This is because in many cases tables may contain figures. Any object that does overlap with the resulting table object is dropped.

\section{Experiments}\label{sec:exps}
We evaluate our model on page object detection datasets. We seek to answer the following questions: (1) how does our end-to-end performance compare against competing object detectors, (2) does our attention-based classification head improve classification results, and (3) how well does our method generalize with respect to document layouts.

\subsection{Experimental Setup}\label{sec:expsetup}
We describe the datasets, metrics, and competing methods we consider.

\subsubsection{Datasets}\label{sec:datasets}
We use two datasets, one that contains images of scientific documents from the Geosciences domain and a benchmark dataset on table, figure, and equation extraction. We describe each dataset in turn.

\newparagraph{Geoscience Paper Dataset} We annotated a collection of 2,466 images from Geoscience papers provided by collaborators on a scientific data extraction project. We divide the pages into a roughly 85:15 train/test split. We provide the class object counts of the train and test partitions in Table~\ref{table:classcounts}.

The dataset is composed of single page images randomly selected from documents, with no pdf metadata provided for each image. As a result, metadata objects cannot be used to segment or classify anything on the page: visual information is all that is provided. Documents are collected from a variety of source conferences and journals with heterogenous layouts. Furthermore, documents can date as far back as 1970, and thus are sometimes not typeset. Some documents are scanned and not rendered, and consequently the quality of the text on the page itself varies greatly.

The Other class is defined as objects in the page that do not contribute scientific content and can be ignored. Wherever possible, annotators were told to create a one to one correspondence between tables/figures and their respective captions, such that groups of tables/figures associated with a single caption are labelled as a single object.

\begin{table*}[t]
\begin{center}
\resizebox{\textwidth}{!}{%
\begin{tabular}{ |c |c |c| c| c| c| c| c|c |c |c| c| c| c|}
\hline
& Body Text & Equation & Figure & Figure Caption & Other & Page Footer & Page Header&Reference Text& Section Header & Table & Table Caption & Total \\
\hline
Train & 6,578 & 1,115 & 1,384 & 1,285 & 359 & 907 & 2,765 & 496 & 3,227 & 767 & 763 & 19,646\\
\hline
Test & 1,258 & 272 & 276 & 274 & 64 & 206 & 533 & 80 & 580 & 172 & 170 & 3,885\\
\hline
\end{tabular}}
\caption{Class counts for Geosciences dataset.}
\label{table:classcounts}
\end{center}
\end{table*}

\newparagraph{ICDAR Dataset} Previous page object detection models generally focus on a single class instead of detecting and classifying all objects on a page. The ICDAR POD 2017 challenge~\footnote{\url{http://u-pat.org/ICDAR2017/program_competitions.php}} was one such competition, with separate challenges for classifying tables, figures, and equations, as well as all three. As we do not have access to the actual test set, we partition the training set into train/test splits, and train using the same optimized hyperparameters as the Geoscience dataset for all models.

\subsubsection{Competing Methods}\label{sec:methods}
We compare our model to state-of-the-art models for page object detection. Page object detection models can be roughly divided into those based off of Faster-RCNN, and those based off of the original R-CNN. We compare to our implementations of both.

\newparagraph{Faster-RCNN} A Resnet-50 convolutional map (with Feature Pyramid Network) of the entire image is produced, which is passed to a region proposal network (RPN). The RPN generates proposals. Our implementation then uses the ROI Align method from Mask-RCNN to map features from the original convolutional map to the proposals (as opposed to the original ROI Pool mechanism from Fast-RCNN). The proposal feature map is then passed to classification and regression heads for final output. This model is derived from a popular implementation of Mask-RCNN~\footnote{\url{https://github.com/matterport/Mask_RCNN}} after dropping the branch that produces masks over proposals. For the RPN, we define anchor sizes of [64, 128, 256, 512, 1024] and aspect ratios of [0.5, 1, 2]. 

\newparagraph{R-CNN} State-of-the-art methods for page object detection generally have performed architecture search over the classification head of the original R-CNN, paired with some document segmentation algorithm to produce proposals. We compare against this class of models by using a ResNet-50 backbone to produce a convolutional map for each cropped and warped proposal. We then pass this to the classification head of Faster-RCNN to produce a class output. No regression over the final box is performed. We implement this model in PyTorch. For this model we use the same grid proposals that we generate for Attentive-RCNN.

We tune the hyperparameters of all three models by partitioning the training set into a training and validation set (80:20 split) and optimizing results over the validation set. We then use these hyperparameters to train over the entire original training set. All backbones are pre-trained on ImageNet using weights available from PyTorch and Tensorflow depending on model implementation.

\subsubsection{Metrics}\label{sec:metrics}
We report standard VOC2012 \cite{everingham_pascal_2010} average precision metrics for our Geoscience dataset experiments, with predictions ranked by classification score. If post-processing is applied and multiple proposals are grouped, the highest score of the group is applied to the newly created output.

From an information retrieval standpoint, average precision does not tell the whole story. Especially with grid segmentation, where all pixels on the page are generally part of a proposal, we are also interested in the overall $F_1$ retrieval metric instead of a summary of the precision-recall curve. We compute $F_1$ using the following quantities:
(1) A \textbf{True Positive} is any proposal with above 0.8 intersection over union (IoU) with a ground truth box, and a correct class assignment. Only one true positive proposal is allowed per ground truth object; (2) A \textbf{False Positive} is any proposal with IoU $> 0.8$ with a ground truth box, but an incorrect class assignment, or a proposal that has 0.7 IoU with a ground truth box for which there is already a match; (3) A \textbf{False Negative} is any ground truth target for which no matched proposal ($>0.8$ IoU) classifies it correctly, and any ground truth target not matched to a proposed region.

For ICDAR, we focus on the classification task: given a matching ground truth target ($>0.8$ IoU), does the model correctly identify its class. We present confusion matrices to show the number of objects retrieved from the dataset.

\subsection{End-to-End Performance}\label{sec:endtoend}
\begin{figure}[t]
\begin{center}
  \includegraphics[width=1.0\columnwidth]{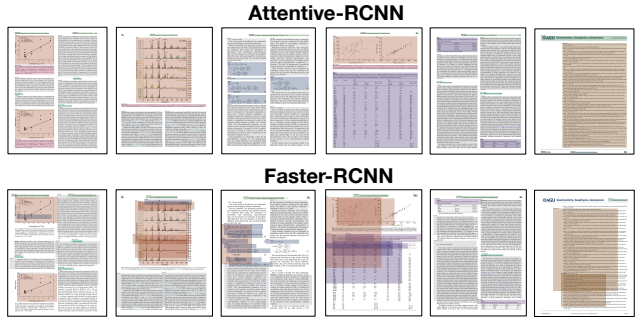}
\end{center}
   \caption{A qualitative comparison between Faster-RCNN and Attentive-RCNN. This figure is zoomable in electronic prints.}
\label{fig:fasterattentive}
\end{figure}

Tables \ref{table:geoscienceap} and \ref{table:geosciencef1} show our AP and F1 results on the Geosciences paper dataset. We see that Attentive-RCNN outperforms the competing R-CNN based models. Visualization of these results can be found in Figure~\ref{fig:fasterattentive}.

\begin{figure}[t]
\begin{center}
  \includegraphics[width=1.0\linewidth]{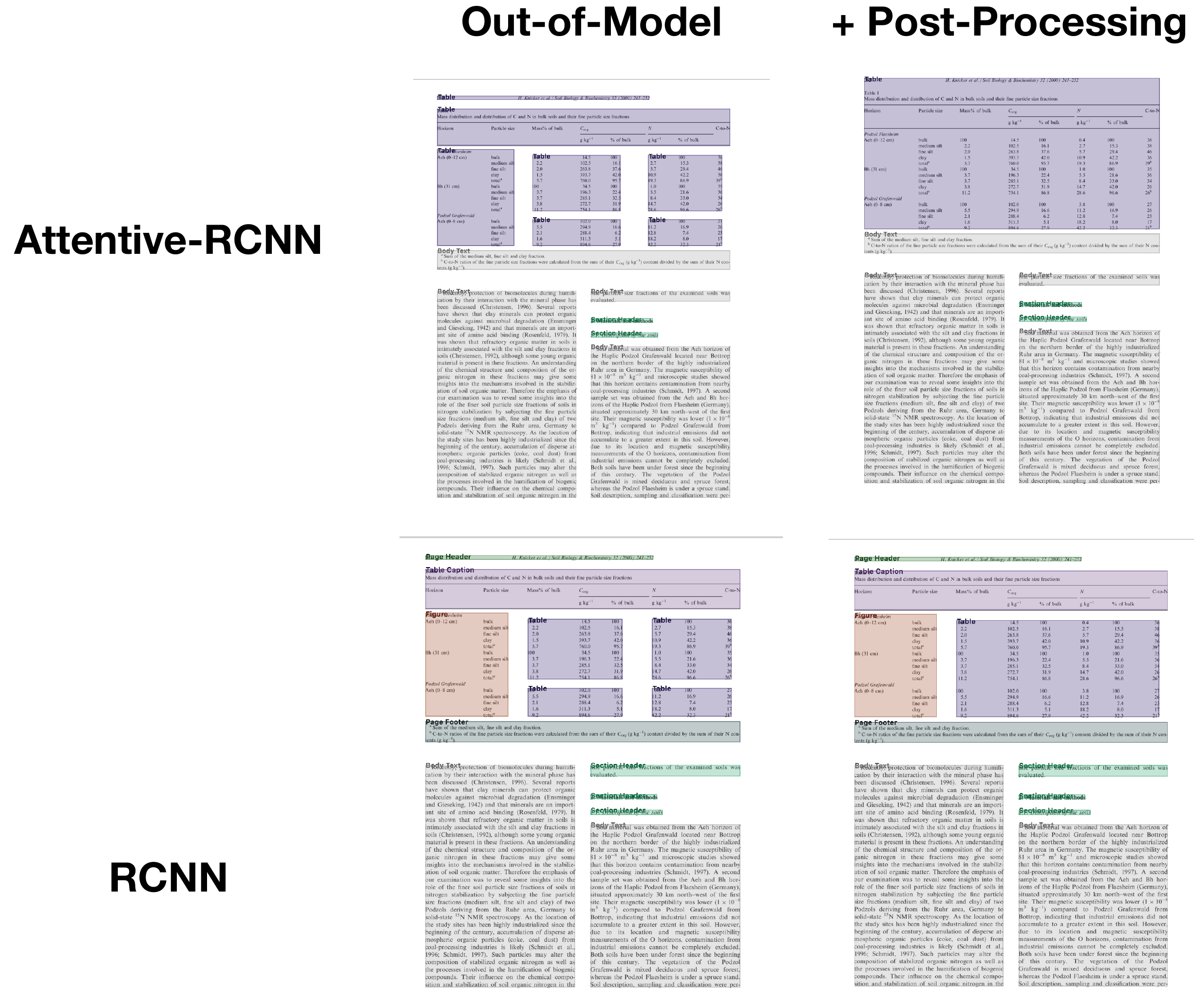}
\end{center}
\caption{Examples demonstrating how attention allows us to obtain higher precision classifications which in turn leads to higher quality extractions after postprocessing.}
\label{fig:postprocessing}
\end{figure}

We see that Attentive-RCNN with postprocessing produces the highest mAP score by $+3\%$ mAP. Both models that use the grid proposals outperform Faster-RCNN significantly, achieving much higher recall of valid objects. When Faster-RCNN is able to recall objects, it tends to classify them correctly, as indicated by the generally high precision across classes (see Table~\ref{table:geosciencef1}). However, because our purpose is information retrieval, failing to retrieve key classes such as equation and table significantly decreases the amount of information that is vital to a document.

Attentive-RCNN generally performs better than R-CNN. For some classes, there are small drops in recall, but those are accompanied by increases in precision. We attribute this behavior to the nature of including context in classification decisions. Despite small drops in recall, increased precision across all classes sans body text implies that context plays a significant role in mitigating class confusion between classes that are not body text. This is key, for example, for making accurate table predictions. Tables that are decomposed as a result of the grid proposal algorithm look like small spans of text. Classifying a small span of text as a table without information about surrounding spans of text is very hard. Coupled with the postprocessing step, we see a significant increase in effectiveness of Attentive-RCNN at detecting tables.

The grouping of tables and figures that occurs as part of the post processing step proves to be important to the retrieval of these object classes (see Figure~\ref{fig:postprocessing}). For Attentive-RCNN, applying the post processing step improves table AP by $+38.3\%$. This large increase implies that decomposed table elements have both low IoU with their ground truth boxes, and high confidence in those decomposed elements (moving them high in the recall ranking list). We see a similar large jump in Attentive-RCNN figure AP ($+39.1\%$). While postprocessing the R-CNN output also increases the AP for these two classes ($+20.9\%$ and $+12.2\%$, respectively), the increase is more significant for Attentive-RCNN because precision over the decomposition is much higher, again highlighting the significance of context in classification.

It is interesting to note that Faster-RCNN produces the highest F1 and mAP for the table caption class. This can be attributed to how proposals break down the page. Most table captions are directly tangent to their tables, with no separating white space. Thus our grid proposal algorithm does not separate the two. For the table captions we do propose, however, we do see higher precision compared to R-CNN. Figure captions, because they are separated from figures by white space, are properly identified by the models that use grid proposals.

Attentive-RCNN successfully utilizes context to improve detection of page object detections. We next examine whether these results can be generalized beyond the domain of Geoscience papers.

\begin{table*}[t]
\begin{center}
\resizebox{\textwidth}{!}{%
\begin{tabular}{ |c |c |c| c| c| c| c| c|c |c |c| c| c| c|}
\hline
& Body Text & Equation & Figure & Figure Caption & Other & Page Footer & Page Header&Reference Text& Section Header & Table & Table Caption & mAP \\
\hline
Faster-RCNN & 0.450 & 0.015 & 0.431 & 0.083 & 0 & 0 & 0.382 & 0.513 & \textbf{0.333} & 0.079 & \textbf{0.137} & 0.220\\
\hline
R-CNN & 0.636 & 0.372 & 0.456 & 0.710 & 0.025 & 0.120 & 0.142 & 0.664 & 0.261 & 0.316 & 0.029 & 0.339\\
+ Postprocessing & 0.615 & 0.371 & 0.665 & \textbf{0.711} & 0.028 & 0.120 & \textbf{0.143} & 0.670 & 0.261 & 0.438 & 0.026 & 0.368\\
\hline
Attentive-RCNN & \textbf{0.676} & \textbf{0.432} & 0.329 & 0.662 & 0.048 & \textbf{0.124} & 0.140 & 0.763 & 0.265 & 0.148 & 0.022 & 0.328\\
+ Postprocessing & 0.652 & \textbf{0.432} & \textbf{0.720} & 0.647 & \textbf{0.049} & \textbf{0.124} & 0.141 & \textbf{0.764} & 0.265 & \textbf{0.531} & 0.022 & \textbf{0.395}\\
\hline
\end{tabular}}
\caption{AP results on the Geoscience paper test set.}
\label{table:geoscienceap}
\end{center}

\end{table*}

\begin{table*}[t]
\begin{center}
\resizebox{\textwidth}{!}{%
\begin{tabular}{|c|c|c|c|c|c|c|c|c|c|c|c|c|c|c|c|}
\hline
& \multicolumn{3}{c|}{Faster-RCNN} & \multicolumn{3}{c|}{R-CNN} & \multicolumn{3}{c|}{R-CNN + Postprocessing} & \multicolumn{3}{c|}{Attentive-RCNN} & \multicolumn{3}{c|}{Attentive-RCNN + Postprocessing} \\
& P & R & $F_1$ & P & R & $F_1$ & P & R & $F_1$ & P & R & $F_1$ & P & R & $F_1$\\
\hline
Body Text & 0.96 & 0.53 & 0.69 & 0.9 & 0.7 & 0.79 & 0.98 & 0.66 & 0.79 & 0.97 & 0.74 & \textbf{0.84} & 0.98 & 0.71 & 0.82\\
\hline
Equation & 1.0 & 0.19 & 0.32 & 0.85 & 0.62 & 0.72 & 0.95 & 0.57 & 0.71 & 0.98 & 0.67 & \textbf{0.79} & 0.98 & 0.67 & \textbf{0.79}\\
\hline
Figure & 0.85 & 0.75 & 0.8 & 0.63 & 0.72 & 0.67 & 0.89 & 0.77 & 0.82 & 0.89 & 0.65 & 0.76 & 0.96 & 0.81 & \textbf{0.88}\\
\hline
Figure Caption & 1.0 & 0.1 & 0.19 & 0.85 & 0.77 & \textbf{0.81} & 0.82 & 0.78 & 0.8 & 0.89 & 0.7 & 0.78 & 0.81 & 0.75 & 0.78\\
\hline
Other & N/A & N/A & N/A & 0.14 & 0.12 & 0.13 & 0.23 & 0.09 & 0.13 & 0.18 & 0.12 & \textbf{0.14} & 0.19 & 0.12 & \textbf{0.14}\\
\hline
Page Footer & N/A & N/A & N/A & 0.76 & 0.5 & \textbf{0.6} & 0.83 & 0.32 & 0.46 & 0.85 & 0.29 & 0.44 & 0.89 & 0.29 & 0.44\\
\hline
Page Header & 0.76 & 0.54 & 0.64 & 0.73 & 0.59 & \textbf{0.65} & 0.84 & 0.39 & 0.53 & 0.85 & 0.38 & 0.52 & 0.87 & 0.38 & 0.53\\
\hline
Reference Text & 0.72 & 0.59 & 0.65 & 0.29 & 0.82 & 0.44 & 0.36 & 0.8 & 0.5 & 0.84 & 0.78 & \textbf{0.81} & 0.84 & 0.78 & \textbf{0.81}\\
\hline
Section Header & 0.91 & 0.47 & 0.62 & 0.8 & 0.52 & \textbf{0.63} & 0.93 & 0.44 & 0.6 & 0.92 & 0.47 & 0.62 & 0.92 & 0.47 & 0.62\\
\hline
Table & 0.98 & 0.18 & 0.3 & 0.48 & 0.61 & 0.54 & 0.68 & 0.65 & 0.67 & 0.8 & 0.55 & 0.66 & 0.86 & 0.66 & \textbf{0.75}\\
\hline
Table Caption & 0.83 & 0.25 & \textbf{0.39} & 0.53 & 0.08 & 0.14 & 0.31 & 0.08 & 0.13 & 0.6 & 0.05 & 0.09 & 0.31 & 0.09 & 0.14\\
\hline
All & 0.86 & 0.4 & 0.55 & 0.71 & 0.58 & 0.64 & 0.80 & 0.52 & 0.63 & 0.87 & 0.51 & 0.65 & 0.87 & 0.54 & \textbf{0.67}\\
\hline
\end{tabular}}
\caption{ F1 Results on Geoscience paper test set. N/A values for a model-class combination mean that the model does not return any predictions for the corresponding class.}
\label{table:geosciencef1}
\end{center}
\end{table*}

\subsection{Generalization to Computer Science papers}
\begin{figure}[t]
\begin{center}
  \includegraphics[width=1.0\columnwidth]{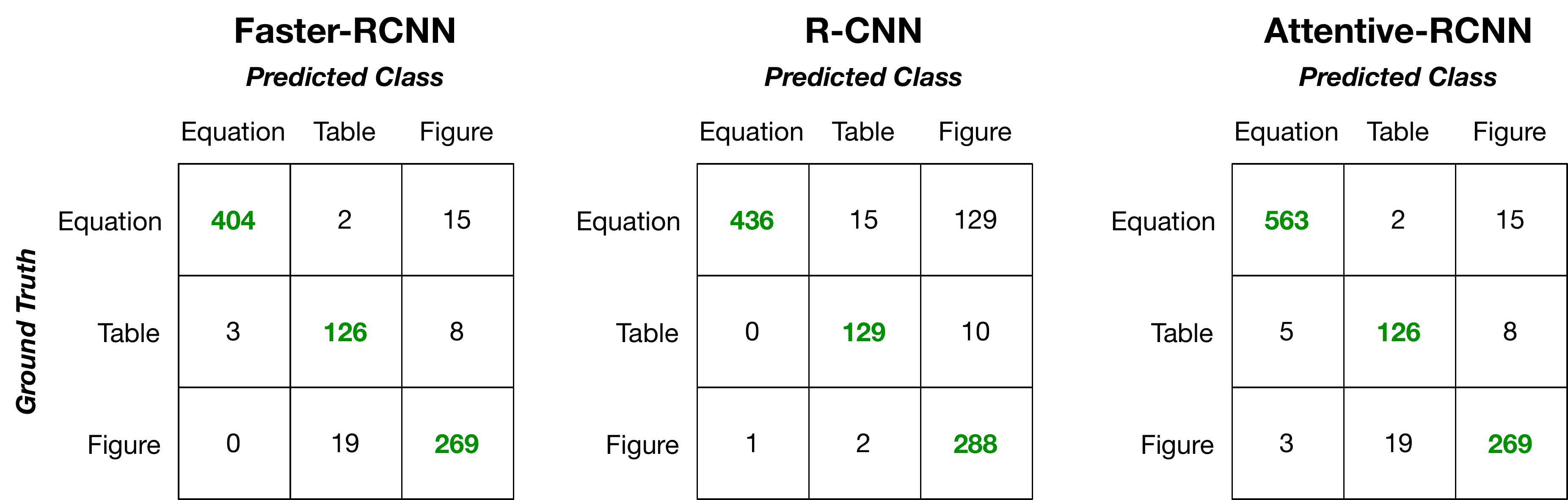}
\end{center}
   \caption{Confusion matrices of results on ICDAR dataset.}
\label{fig:confusion}
\end{figure}

We now evaluate whether our model can be applied to another document domain with a different standard for document layouts. The ICDAR POD 2017 dataset is composed of recent Computer Science papers, with images that are distinct from the Geosciences datasets. Each model is trained in the exact same way as the Geosciences dataset. For R-CNN and Attentive-RCNN, proposals which have no mapping to a ground truth target are not trained on. The test set contains 559 figures, 187 tables, and 1,077 equations. Confusion matrices showing results are shown in Figure \ref{fig:confusion}. Attentive-RCNN retrieves 127 more correct equations than R-CNN by using context to successfully handle equation/figure class confusion. Faster-RCNN and Attentive-RCNN correctly classify high IoU regions, but Attentive-RCNN is able to retrieve more high IoU regions because the grid proposals are of higher quality than Faster-RCNN's proposals.

\section{Conclusion}\label{sec:conclusions}
We show that context is important in classification for the page object detection task. Our attention mechanism, paired with region embeddings, enables frameworks that algorithmically produce proposals to introduce contextual information from the regions surrounding proposals. While we focus on information extraction from scientific documents here, the attention framework we have proposed can be extended to natural scene images, which we leave as a future research direction.

\section{Acknowledgements}
We gratefully acknowledge the support of DARPA under No. HR00111990013  (ASKE). The U.S. Government is authorized to reproduce and distribute reprints for Governmental purposes notwithstanding any copyright notation thereon. Any opinions, findings, and conclusions or recommendations expressed in this material are those of the authors and do not necessarily reflect the views, policies, or endorsements, either expressed or implied of DARPA or the U.S. Government. 

{\small
\bibliographystyle{ieee}
\bibliography{egbib}
}

\end{document}